\documentclass[letterpaper, 10 pt, conference]{ieeeconf}  

\IEEEoverridecommandlockouts                              

\usepackage{graphicx} 
\usepackage{caption}
\usepackage{hyperref}
\usepackage{xcolor}

\title{\LARGE \bf
A Saliency based Feature Fusion Model for EEG Emotion Estimation}

\author{Victor Delvigne$^{1,2,*}$, Antoine Facchini$^{2}$, Hazem Wannous$^{2}$, Thierry Dutoit$^{1}$ \\Laurence Ris$^{3}$ and Jean-Philippe Vandeborre$^{2}$
\thanks{$^{1}$ ISIA Lab, Faculty of Engineering, \textit{University of Mons}, Belgium;}%
\thanks{$^{2}$ IMT Nord Europe, \textit{CRIStAL UMR CNRS 9189}, France;}%
\thanks{$^{3}$ Neuroscience Lab, Faculty of Medecine and Pharmacy, \textit{University of Mons}, Belgium}%
\thanks{$^{*}$Corresponding author {\tt victor.delvigne@umons.ac.be}}%
}

\begin{document}

\maketitle
\thispagestyle{empty}
\pagestyle{empty}

\begin{abstract}
Among the different modalities to assess emotion, electroencephalogram (EEG), representing the electrical brain activity, achieved motivating results over the last decade. Emotion estimation from EEG could help in the diagnosis or rehabilitation of certain diseases. In this paper, we propose a dual model considering two different representations of EEG feature maps: 1) a sequential based representation of EEG band power, 2) an image-based representation of the feature vectors. We also propose an innovative method to combine the information based on a saliency analysis of the image-based model to promote joint learning of both model parts. The model has been evaluated on four publicly available datasets: SEED-IV, SEED, DEAP and MPED. The achieved results outperform results from state-of-the-art approaches for three of the proposed datasets with a lower standard deviation that reflects higher stability. For sake of reproducibility, the codes and models proposed in this paper are available at \url{https://github.com/VDelv/Emotion-EEG}.
\end{abstract}

\section{INTRODUCTION}
\label{sec:intro}
Emotion estimation is a trending topic in various research applications. Advances in this field could help to have a better understanding of language processing and non-verbal communication in the context of Human-Computer-Interaction (HCI). Although most of the proposed works consider voice/sound, video, images or text, recent works have shown that the emotional state can also be predicted from biomedical signals. An example is the MPED corpus \cite{song_mped_2019} presenting a dataset composed of physiological recordings, i.e. EEG, electrocardiogram (ECG), respiration and galvanic skin response (GSR), with an initial benchmark aiming to estimate emotional state are presented. Furthermore, other works have shown that eye-tracking signals can also be considered to estimate emotion \cite{zheng_multimodal_2014}.

During the last decade, Artificial Intelligence (AI) and specially Machine-Learning (ML) based algorithms have known an increase in interest. The reasons are varied: higher robustness, higher accuracy, technological democratisation and the growing simplicity of their development. Through the scientific community, ML-based algorithms are used in a wide range of technologies and domains including medical field with for instance: diseases and infections detection from physiological recordings \cite{hatt_first_2018}, understanding of the human genomes \cite{li_genome-wide_2018}, human behaviour prediction \cite{phan_deep_2016} or drugs discovery \cite{pastur-romay_deep_2016}. 

On another side, the recent researches on Brain-Computer Interfaces (BCI) have led to an increase in their use in innovative applications and projects. BCI aims to create a connection between computers and users' brains through physiological signals, e.g. EEG, Magnetoencephalogram (MEG), Magnetic Resonance Images (MRI). This connection is invasive/non-invasive; open-loop (e.g. recording signals during a specific task on a computer) or closed-loop (e.g. video game evolving with the brain activity) depending on the considered application.

In the context of emotion estimation, an increase of research projects proposing BCI application have been noted. As reported by Craik et al. \cite{craik_deep_2019}, one paper out of 6 considering the use of Deep Learning (DL) algorithm for EEG is dedicated to emotion estimation tasks. The existing works present datasets to assess emotional state \cite{zheng_investigating_2015, zheng_emotionmeter_2019, song_mped_2019, koelstra_deap_2012} and different models often based on DL algorithms to retrieve the emotional state from EEG signals \cite{li_novel_2020, song_eeg_2018, zong_novel_2018, zhong_eeg-based_2020}. The proposed methodologies consider EEG under different forms: e.g. graph representation of EEG signals \cite{song_eeg_2018, zhong_eeg-based_2020}, vectors separated between each hemispheres \cite{li_novel_2020, zong_novel_2018} and even images based representation \cite{van2021insights}.

One of the major concerns for the use of ML algorithms in medical applications and especially for EEG processing is the explainability of models. Recent works \cite{li_perils_2021} have shown that biases may exist in several works considering ML-based processing of EEG. In this context, it is important to consider interpretable models. Different approaches have been designed for this purpose. For instance, the signal can be projected in a latent space to have a better representation of the signals \cite{patel_latent_2013} to help in the classification. Another approach \cite{selvaraju2017grad} consists in visualizing the elements (i.e. pixels of an image, point of a graph or element of an array) used by the model to make a decision.

In this context, a framework aiming to estimate emotion from EEG is presented. This framework is composed of two parallel modules: (1) a higher-level network considering an image-inspired representation of EEG to benefit from advantages of computer vision models; (2) a lower-level network considering each electrode contribution through an array representation of EEG. The contributions of the two networks are then combined to estimate the corresponding emotion state from a given EEG trial.

\section{METHOD}


As in previous works, the proposed framework follows a general pipeline to estimate emotion from EEG signals. In this section, each step composing this pipeline is presented.

\subsection{Feature extraction}

Let's consider an EEG sample $\in \mathrm{R} ^{n_{channels} \times n_{samples}}$ with $n_{channels}$ the number of electrodes on the EEG headset and $n_{samples} = duration * f_{sampling}$ the sample size for each trial. It is possible to manually extract $n_{features}$ for each EEG segments considered as time series to have a representation of EEG in a smaller subspace $\in \mathrm{R}^{n_{channels} \times n_{features}}$. The considered features may represent temporal, spatial (i.e. related to electrodes location on the scalp) or frequential (i.e. feature related to the contribution of different frequential bands) information from signals. Due to the difficulty to characterize raw EEG signals and their trend to be affected by noise (e.g. electrical noise and muscles given their close frequency to EEG \cite{muthukumaraswamy_high-frequency_2013}), a majority of dataset proposed denoised EEG and pre-extracted features \cite{zheng_investigating_2015, song_mped_2019, zheng_emotionmeter_2019}. Among the most commonly considered feature extraction methods for emotion estimation, two have been kept: 1) the power spectral density (PSD) representing the contribution of each frequential band in the EEG signal, 2) the differential entropy (DE) reflecting the temporal evolution of EEG segments \cite{duan_differential_2013}. The feature extraction methods can be considered separately or combined by computing DE on EEG signals filtered at specific frequencies. The choice of these feature extraction methods is motivated by their encouraging results for emotion estimation \cite{van2021insights, duan_differential_2013, zheng_investigating_2015, ahern_differential_1985}.

From the array representation, it is also possible to consider spatial information through more visual representation image-based EEG feature maps \cite{bashivan_learning_2015}. Given the location of the electrodes in a 3D frame (i.e. cartesian coordinate of the position on the scalp), it is possible to consider an azimuthal projection to represent their locations in 2D. Finally, after assigning the feature values to each electrode in the 2D discrete representation an image is created by interpolating the values in the two projected dimensions. Finally, the constituted images will have the following shape $[n_{features} \times h \times w]$ with the height $h$ and width $w$ taken arbitrarily.

\begin{figure*}
    \centering
    \includegraphics[width=0.99\textwidth]{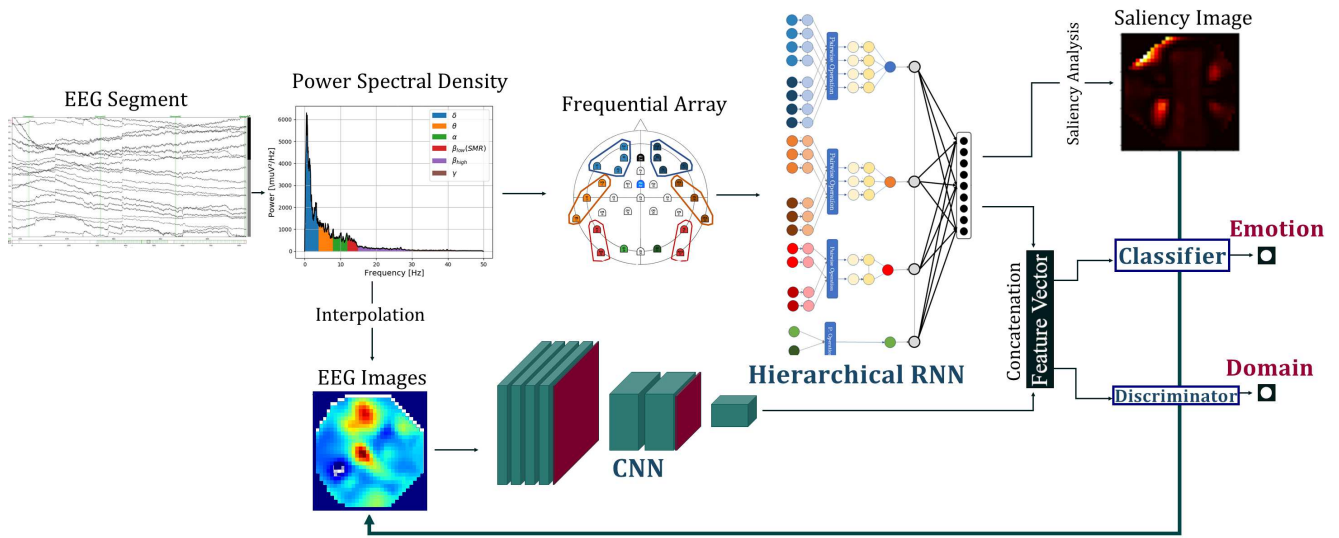}
    \caption{Overview of the framework composed of the feature extraction and represenation (left part), the H-RNN (top part), the CNN (bottom part) and the emotion and domain classification (right part). The saliency maps extracted by the H-RNN are used to weight input images of the CNN.}
    \label{fig:Model}
\end{figure*}

\subsection{Images estimation}
From the image representation of EEG signals $\in \mathrm{R}^{n_{feat} \times h \times w}$, two-family of models, initially dedicated to image processing, have been considered: convolutional neural networks (CNN) and capsule networks. 
 
The CNN approach consists of a VGG-inspired model \cite{simonyan_very_2015} with an architecture composed of three modules each of them respectively composed of 4, 2 and 1 convolutional layers followed by batch normalization layers. Each module is separated by a max-pooling layer. 

The capsule-based model consists of the general architecture presented by Sabour et al. \cite{sabour_dynamic_2017}. One of the advantages of this network is its ability to extract the spatial information between elements composing an image, e.g. automatically taking into account the nose or mouth position for face detection systems. In our method, the initial dynamic routing between capsules has been adapted to match EEG images. The consideration of this novel approach allows us to study the spatial relevance among EEG images. Moreover, the use of capsule networks remains also interesting for their lower computational cost compared to other CNN architectures during inference. 

\subsection{Array estimation}
On the other hand, a more understandable approach compared to the previously presented models has also been considered. This last consists of a recurrent neural network (RNN) composed of the succession of hierarchical sub-networks, each of these networks aiming to extract information at different levels: electrodes, physiological regions or hemispheres based relationships. This model has been inspired by the hierarchical RNN (H-RNN) \cite{yong_du_hierarchical_2015} aiming to classify motor movements from skeleton recordings at different locations and levels, e.g. fingers, hands, arms, trunk.

The network aims at considering the contribution of each element at each spatial level: relative electrodes positions, physiologically defined electrodes regions and hemispheres. A representation of the framework with the H-RNN is shown in Figure \ref{fig:Model}. During the training, it is possible to compute the activation of each level and then to measure the contribution of each element composing the feature array to estimate the emotional state. 

More, RNN has shown motivating results on various public EEG datasets and especially for emotion estimation \cite{li_novel_2020}.

\subsection{Merging models methodology}
From the results provided by the models presented above, a combination of these last has been considered to merge their strengths and improve the classification accuracy. Different methodologies can be considered depending on the fusion location of the models: 1) output fusion considering a linear combination of both estimated classes; 2) feature fusion consisting in concatenating of the features vectors extracted by the two DL models.

To promote the synchronised learning of both networks and help transfer learning between modules, a novel approach based on saliency analysis has been considered. This last consists of estimating the most salient electrodes/regions of the scalp to make the estimation for the RNN and to integrate this information during the training of the image-based network. The feature vector saliency from the RNN is computed as $ Saliency = \left\| \frac{\partial Class}{\partial X} \right\|$. These vectors measure the importance of each electrodes' feature to estimate emotion. From the saliency vector, an image representation is considered and used to weigh the image representation of the feature vector. This process aims to concentrate the learning around the most important region of the EEG.

\section{EXPERIMENT}

In this section, the considered datasets for training the proposed model are presented, then its implementation details are presented and an adversarial training methodology as a domain classification is proposed.

\subsection{Datasets}
We consider four different datasets to evaluate our approach. The choice of working with four different datasets, despite optimizing only one of them, is done to promote a general approach working on different participants of different backgrounds. 
\begin{itemize}
    \item SEED IV \cite{zheng_emotionmeter_2019} contains EEG recordings of 15 different participants spread over 3 sessions each consisting of 24 trials. One trial consists of EEG recording during video clips promoting several emotions, during which several physiological signals are recorded (eye positions and EEG). In this dataset, four emotion classes have been considered: \textit{happy, sad, neutral and fear}.
    \item SEED \cite{zheng_investigating_2015} is composed of recordings from 3 sessions repeated for 15 participants, each session is composed of 15 trials. The experimental setup consists of the recording of EEG during video promoting specific emotion. In the SEED dataset, emotions have been separated into three more general classes: \textit{positive, negative and neutral}.
    \item MPED \cite{song_mped_2019} contains the recordings of several physiological signals (EEG, ECG, GSR) for a total of 30 participants. Similarly, the recordings have been made during video clips. The promoted emotion during the videos have been separated into seven classes: \textit{joy, funny, anger, disgust, fear, sad and neutrality}.
    \item DEAP \cite{koelstra_deap_2012} is a multi-modal dataset composed of EEG, electromyogram (EMG) and electrooculogram (EOG) recordings of 32 participants. During the dataset creation, it has been asked to participants to look at the video and to self-assess emotion state according to three dimensions: arousal (i.e. from excitation to disinterest), valence (i.e. from pleasant to unpleasant) and dominance (i.e. the ability to control the feelings from weak to empowered) considered as the labels of the physiological recordings. 
\end{itemize}

Each dataset has been recorded with biomedical EEG headsets following the 10/20 electrodes placement of the 62 electrodes for \cite{zheng_emotionmeter_2019, zheng_investigating_2015, song_mped_2019} and 32 electrodes for \cite{koelstra_deap_2012} (proposed model has been adapted to fit the number of considered electrodes). Moreover, the provided feature vectors have been considered in this experiment: DE in five frequency bands ($\delta, \theta, \alpha, \beta$ and $\gamma$ bands) in \cite{zheng_emotionmeter_2019, zheng_investigating_2015}. For \cite{song_mped_2019, koelstra_deap_2012} PSD has been considered with the same frequency bands limit and methodologies as in the original paper \cite{koelstra_deap_2012}.

\subsection{Domain classification}
It has been noted that the recent advance in DL leads to novel training methodology increasing estimation accuracy. One of them aims at transferring the information from a dataset to another \cite{ganin_unsupervised_2015}. This method consists in training the network with the original dataset similarly to regular cross-validation methods with an additional step. During this step, both datasets are used but their labels are hidden (thus considered as unknown by the network) and the goal for the network during this step is to estimate the source dataset of the considered element. This additional step allows to add knowledge from other datasets and also to extract information from input regardless of their source. 

In the context of EEG processing, a similar approach has been proposed \cite{lan2018domain} and consider the training and validation as the two different datasets. The methodology remains valid since the labels are not used during the domain classification. The domain classification helps to improve the cross-validation accuracy by promoting more general processing of information by the DL model \cite{lan2018domain}.

\subsection{Implementation details}
In image-based representation, the number of channels for the three CNN sub-modules has been respectively set to 16, 64 and 128. Moreover, we consider $32 \times 32$ images to consider square shape images that better fit with convolution. For the other representation, the hidden dimension of the RNN has been set to 32. During the training phase, a cross-entropy loss has been considered, moreover, adam optimizer with an adaptative learning rate has been considered: $l_r = 10{-3}$ and weight decay, $w_d = 10^{-8}$. The number of epochs has been fixed to 150, if the loss was not evolving favourably during 5 epochs the training was stopped. All the models have been implemented with Pytorch library and were trained on one 24 GB Nvidia Titan RTX GPU. For the sake of reproducibility, the model architecture is freely available online.\footnote{\url{https://github.com/VDelv/Emotion-EEG}}

\section{RESULTS and DISCUSSION}

\begin{table}[t!]
    \centering
    \begin{tabular}{l | c}
        Model & Accuracy ($\mu/\sigma$) [\%] \\
        \hline
        SVM \cite{li_novel_2020}& 31.46/9.20 \\
        DGCNN \cite{song_eeg_2018} & 52.82/9.83 \\
        BiDANN \cite{zong_novel_2018} & 65.59/10.39 \\
        BiHDM \cite{li_novel_2020}& 69.03/8.66 \\
        RGNN \cite{zhong_eeg-based_2020} & 73.84/8.02 \\
        H-RNN$^*$ & 63.56/5.74 \\
        Image EEG (CNN)$^*$ & 64.97/6.15 \\
        Image EEG (Capsule)$^*$ & 60.83/8.53 \\
        Output Fusion$^*$ & 69.34/5.14 \\
        Feature Fusion$^*$ & 71.48/5.02\\
        \textbf{Saliency Fusion}$^*$ & \textbf{74.42/4.76} \\
    \end{tabular}
    \caption{Leave One Subject Out cross-validation accuracy for emotion estimation on SEED-IV dataset \cite{zheng_emotionmeter_2019}.$^*$ denotes the results obtained from our model experiments.}
    \label{tab:CompMod}
\end{table}

To assess the results provided by the proposed models, we have first considered the training and validation of the different approaches proposed above on the SEED-IV dataset \cite{zheng_emotionmeter_2019}, this last being the bigger. Then, the best approaches have been compared to the state-of-the-art models on the three other datasets \cite{zheng_investigating_2015, song_mped_2019, koelstra_deap_2012}.

The considered metric for model evaluation is the Leave One Subject Out (LOSO) cross-validation accuracy. With this evaluation methodology, all the subjects except one are used to train the model and the evaluation is made on the remaining one. This evaluation is repeated for each participant and the corresponding mean and standard deviation are computed. The LOSO cross-validation accuracy has been chosen to assess the model ability to generalise to participants never met previously. EEG signals being very person-specific \cite{li_perils_2021}, a large gap is often noted for the cross-validation accuracy between participant dependant and participant independent (i.e. LOSO). Nevertheless, BCI applications are supposed to be directly used on the participant in real-life, i.e. their signals are not used during the training of the DL model. It was thus decided to consider LOSO cross-validation to monitor the model ability to generalise.  

As reported in Table \ref{tab:CompMod}, all the described approaches previously mentioned have been trained on the SEED-IV \cite{zheng_emotionmeter_2019} dataset. As shown, the best results for the primary study (standalone approach without models combination) were found for the image-based approach with CNN. Besides, lower results were also noted for the hierarchical RNN with higher stability among participants (i.e. lower standard deviation). For this reason, it has been chosen to consider the combination of these two models to merge their advantages. 

Three different approaches have been used to merge the models: concatenating the output estimation (Output Fusion), concatenating feature vectors with saliency extraction from the low-level model (Saliency Fusion) and without saliency (Feature Fusion). As reported in Table \ref{tab:CompMod}, the merged models present better results, especially the saliency-based combination that exceeds the results from state-of-the-art methods but also our previous experiments. Another advantage of the saliency-based feature fusion is its low standard deviation compared to other models. This expresses the fact that saliency-based fusion presents similar results independently of the participants. The improvements provided by the saliency-based approach are explained by this dual methodology considering in parallel the sequential information provided by the H-RNN and the more general region activation highlighted by the CNN. Furthermore, focusing the training on the specific region of the image-based EEG with a raw attention mechanism presents better performance than a simple concatenation.

\begin{table}[b]
    \centering
    \caption{}
    \begin{tabular}{l c c c c}
        Dataset & SEED-IV & SEED & DEAP & MPED \\
        \hline
        BiHDM \cite{li_novel_2020}  & 69.03/8.6 & \textbf{85.40/7.5} & - & 28.27/4.9 \\
        RGNN \cite{zhong_eeg-based_2020} & 73.84/8.1 & 85.30/6.7 & - & - \\
        RODAN \cite{rodan_lew_eeg-based_2020} & 60.75/10.4 & - & 56.60/3.5 & - \\
        \textbf{Saliency}$^*$ & \textbf{74.42/4.8} & 84.11/2.9 & \textbf{78.47/4.9} & \textbf{32/4.7} \\ 
    \end{tabular}
    \captionsetup{labelformat=empty}
    \caption{Leave One Subject Out cross-validation accuracy for emotion estimation on several datasets: SEEED-IV \cite{zheng_emotionmeter_2019}, SEEED \cite{zheng_investigating_2015}, MPED \cite{song_mped_2019} and DEAP \cite{koelstra_deap_2012}.$^*$ denotes the results obtained from our model experiments.}
    \label{tab:CompDat}
\end{table}

In Table \ref{tab:CompDat}, a comparison of the results of two of the previously mentioned models are presented with state-of-the-art models for emotion estimation from EEG. As shown, our approaches present the best results for some datasets and remain on the same scale for other datasets that proved their ability to estimate emotion in various cases.  More, our approach presents better results than \cite{van2021insights}, however, the exact scores being not given in their paper, our results have been compared to their results in their presented Figures. As seen in Table \ref{tab:CompDat}, the results obtained by our saliency-based approach exceed those obtained by previous works. Although this last may seem only slightly better than previous works, it is important to note that the proposed approach is able, at least, to achieve a comparable subject independent cross-validation accuracy as previous works. The purpose of the proposed method is to present a general model instead of a finely tuned approach working only in a specific context.

\section{CONCLUSION}
In this paper, we proposed a novel framework aiming to estimate emotion from EEG. The proposed model is composed of a dual approach considering the spatial relationship between EEG channels through a hierarchical RNN and a DL representation through CNN. The proposed method shows encouraging results on three datasets. In further work, it could be interesting to consider novel ML models with different representations (e.g. graph neural network) and more understandable approaches with other feature extraction methods.

In the next few years, emotion estimation from EEG could be used in various applications covering several fields, e.g. entertainment or medical domain.


\section*{ACKNOWLEDGMENT}
Research jointly supported by the University of Mons and Institut Mines-Télécom Nord Europe. The content is solely the responsibility of the authors. Any opinions, findings, conclusions or recommendations expressed in this material are those of the author(s). This work has been made in collaboration with the Centre de Recherche et de Formation Interdisciplinaire en Psychophysiologie et Electrophysiologie de la Cognition (CiPsE). The authors would like to thank Nathan Hubens and Luca La Fisca for their collaboration.

\bibliographystyle{plain}
\bibliography{Bibliography}

\end{document}